\documentclass{svproc}

\usepackage{url}
\usepackage{amsmath,amssymb,amsfonts}
\usepackage{algorithmic}
\usepackage[ruled,vlined,boxed,linesnumbered]{algorithm2e}
\usepackage{graphicx}
\usepackage{textcomp}
\usepackage{booktabs}
\usepackage{bm}
\usepackage{hyperref}
\usepackage{multirow}
\usepackage{multicol}
\usepackage{makecell}
\usepackage{array}

\usepackage{enumerate}
\usepackage{enumitem}

\newcommand{\PreserveBackslash}[1]{\let\temp=\\#1\let\\=\temp}
\newcolumntype{C}[1]{>{\PreserveBackslash\centering}p{#1}}
\newcolumntype{R}[1]{>{\PreserveBackslash\raggedleft}p{#1}}
\newcolumntype{L}[1]{>{\PreserveBackslash\raggedright}p{#1}}

\newcommand{\argmin}{\mathop{\mathrm{argmin}}}
\newcommand{\argmax}{\mathop{\mathrm{argmax}}}

\begin{document}
\mainmatter

\title{FedBiCross: Personalized One-Shot Federated Learning on Medical Images}

\titlerunning{FedBiCross}

\author{Yuexuan Xia\inst{1}\thanks{Equal contribution.} \and 
Yinghao Zhang\inst{2}\footnotemark[1] \and 
Yalin Liu\inst{2}\thanks{Corresponding author.} \and 
Hong-Ning Dai\inst{3} \and 
Yong Xia\inst{1}\footnotemark[2]}

\authorrunning{Yuexuan Xia et al.}

\tocauthor{Yuexuan Xia, Yinghao Zhang, Yalin Liu, Hong-Ning Dai, Yong Xia}

\institute{School of Computer Science and Engineering, Northwestern Polytechnical University, China\\
\email{xiayuexuan@mail.nwpu.edu.cn, yxia@nwpu.edu.cn}
\and
School of Science and Technology, Hong Kong Metropolitan University, Hong Kong\\
\email{zhangyi@hkmu.edu.hk, ylliu@hkmu.edu.hk}
\and
Department of Computer Science, Hong Kong Baptist University, Hong Kong\\
\email{henrydai@comp.hkbu.edu.hk}}

\maketitle

\begin{abstract}
Data-free knowledge distillation-based one-shot federated learning (OSFL) trains a model in a single communication round without sharing raw data, making OSFL attractive for privacy-sensitive medical applications. However, existing methods aggregate predictions from all clients to form a global teacher. Under non-IID data, conflicting predictions dilute each other during averaging, yielding less informative soft labels that weaken distillation. We propose FedBiCross, a personalized OSFL framework with three stages: (1) clustering clients by model output similarity to form coherent sub-ensembles, (2) bi-level cross-cluster optimization that learns adaptive weights to selectively leverage beneficial cross-cluster knowledge while suppressing negative transfer, and (3) personalized distillation for client-specific adaptation. Experiments on four medical image datasets demonstrate that FedBiCross consistently outperforms state-of-the-art baselines across different non-IID degrees.
\end{abstract}

\section{Introduction}
\label{sec:intro}

By sharing medical images across institutions, such as hospitals, clinics, and imaging centers, vision-based artificial intelligence models can be trained on large-scale datasets, thus improving their performance in medical applications~\cite{guan2024federated}. Nonetheless, the sharing of across-institutional data is strictly restricted by privacy regulations~\cite{kaissis2020secure}. To address this issue, Federated learning (FL) provides a promising solution for high-privacy data sharing and collaborative model training~\cite{mcmahan2017communication}. In FL, participating institutions are deemed multiple \emph{clients} to collaboratively train a global model on a server while keeping trained data in local systems.

However, conventional FL needs timely and numerous communication rounds between clients and the server, hard to achieve in medical networks because of regulated data transfer procedures~\cite{guha2019one} and communicational resource constraints \cite{wang2025discovering}. For instance, institutional review boards may grant data access only within narrow time windows, and each additional round of model exchange enlarges the attack surface for inference attacks~\cite{guha2019one} and leads to large communication overhead. 
One-shot federated learning (OSFL)~\cite{guha2019one} addresses these two challenges by restricting communication to a single round. Among various OSFL strategies, data-free knowledge distillation (DFKD) is particularly attractive because it avoids transmitting any data-dependent surrogates~\cite{zhang2022dense,kang2023one,dai2024enhancing}. 
In DFKD, each client uploads its locally trained model once, and the server synthesizes images from random noise via DeepInversion and trains a student model using the averaged predictions of all uploaded client models as soft labels~\cite{yin2020dreaming}. 

Though the benefits of DFKD, it faces a critical limitation in handling non-IID data in medical networks, i.e, client data of different institutions varies in patient population, label proportions, imaging protocols, and disease prevalence~\cite{yan2025fedvck}. Existing DFKD methods aggregate predictions from \emph{all} clients to form a single global teacher, but each client model is biased toward its locally frequent categories. For the same synthetic input, different clients produce conflicting predictions, and naively averaging them yields near-uniform soft labels that carry little discriminative information for distillation. Moreover, even if the soft label quality were improved, a single global model cannot adequately serve all clients under heterogeneous distributions. A natural remedy is to cluster similar clients into sub-ensembles, but this in turn raises the questions of how to exploit knowledge across clusters without harmful interference, and how to bridge the remaining gap between a cluster-level model and each client's individual distribution. Therefore, we summarize three significant challenges:
\begin{enumerate}[leftmargin=*,align=left]
    \item[\emph{Challenge 1:}]~How to form coherent ensemble teachers that produce informative soft labels under heterogeneous client data distributions?
    \item[\emph{Challenge 2:}]~How to selectively leverage cross-cluster knowledge while suppressing negative transfer?
    \item[\emph{Challenge 3:}]~How to adapt cluster-level models to individual clients whose distributions may still differ within the same cluster?
\end{enumerate}

While personalized FL methods such as local fine-tuning~\cite{oh2021fedbabu}, 
and clustering-based training~\cite{briggs2020federated,ghosh2020efficient,vahidian2023efficient} address non-IID heterogeneity in conventional FL, they rely on iterative communication and are inapplicable to the one-shot setting. To address the above challenges within a single communication round, we propose \textbf{FedBiCross}, a personalized one-shot FL framework. Our key insight is that clients serving similar patient populations tend to produce consistent predictions, so grouping them into coherent sub-ensembles can yield peaked and confident soft labels for effective distillation. This motivates a bi-level optimization scheme that learns adaptive cross-cluster weights by evaluating their contribution to target-cluster performance. Finally, since clients within the same cluster may still differ in imaging equipment, patient demographics, or clinical protocols, we introduce personalized distillation that adapts cluster models to individual clients. Our main contributions are:

\begin{itemize}
    \item We propose a client clustering strategy based on model output similarity that groups clients into coherent sub-ensembles, producing more peaked and informative soft labels for effective distillation under non-IID settings.
    \item We introduce a bi-level cross-cluster optimization framework that automatically learns adaptive weights to selectively leverage beneficial cross-cluster knowledge while suppressing negative transfer.
    \item We develop a personalized distillation stage that adapts cluster-level models to individual clients, capturing client-specific patterns while preserving generalizable cross-cluster knowledge.
\end{itemize}

\section{Methodology}
\label{sec:method}

\begin{figure}[!ht]
\centerline{\includegraphics[width=\textwidth]{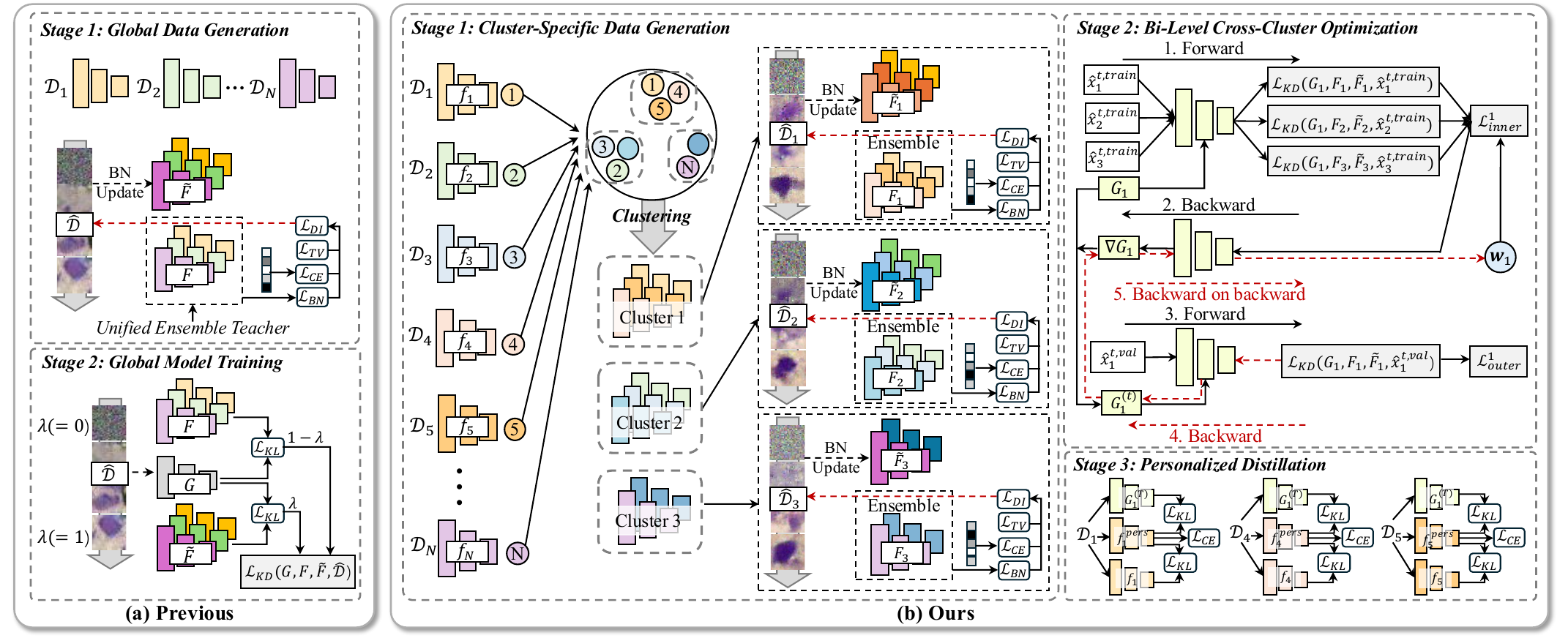}}
\caption{Overview of FedBiCross. (a) Previous methods construct a unified 
ensemble teacher from all clients. (b) Our three-stage approach: client 
clustering, bi-level cross-cluster optimization with online synthesis, 
and personalized distillation.}
\label{pipeline}
\end{figure}

We consider a federated learning system with $N$ clients and a central server. 
Each client $i$ holds a private local medical image dataset $\mathcal{D}_i$ and has trained a 
local model $f_i$ on it. Our goal is to produce personalized models 
$\{f_i^{\mathrm{pers}}\}_{i=1}^N$ for all clients within a single communication 
round: each client uploads $f_i$ to the server once, and the server returns a 
personalized model without accessing any raw data. We assume the server possesses 
abundant computational resources, making it practical to perform intensive 
server-side optimization in exchange for minimal communication. To this end, we propose \textbf{FedBiCross}, a framework for personalized OSFL. Figure~\ref{pipeline} illustrates the overall framework and Algorithm~\ref{alg:fedbicross} summarizes the procedure. FedBiCross consists of three stages: (1) \textit{client clustering}, (2) \textit{online synthesis with bi-level cross-cluster optimization}, and (3) \textit{personalized distillation} to fine-tune models for each client.

\subsection{Stage 1: Client Clustering}

\subsubsection{Client Clustering via Similarity.}
To obtain confident supervision, we group clients whose models produce similar predictions into coherent sub-ensembles. Since our goal is to align predictions within each ensemble teacher, we measure client similarity directly in the output space (an approach shown effective by FLIS~\cite{morafah2023flis}) rather than in the parameter or feature space, where different parameterizations can yield identical predictions. 

Specifically, the server generates a set of $M$ random noise images $\{\bm{z}_m\}_{m=1}^M$ and computes predictions from each client model: $\bm{p}_i = [f_i(\bm{z}_1), \ldots, f_i(\bm{z}_M)] \in \mathbb{R}^{M \times C}$, where $C$ is the number of classes and $f_i$ is the local model. We apply $K$-means clustering on the prediction matrices: $\min_{\{\mathcal{C}_k\}, \{\bm{c}_k\}} \sum_{k=1}^{K} \sum_{i \in \mathcal{C}_k} \left\| \bm{p}_i - \bm{c}_k \right\|_F^2$, where $\|\cdot\|_F$ denotes the Frobenius norm, $\{\mathcal{C}_k\}_{k=1}^K$ are the resulting clusters, and $\bm{c}_k$ denotes the centroid of cluster $k$. We use randomly generated images as probe inputs because (i) the data-free setting precludes access to real data on the server, and (ii) random noise naturally aligns with the starting point of DeepInversion (Eq.~\ref{eq:synthesis_update}), ensuring that the clustering reflects the same input regime under which the ensemble teachers will operate during synthesis.

\paragraph{Automatic Selection of $K$.}
A critical question is how to determine the number of clusters $K$ without manual tuning. We adopt the Silhouette Score as an automatic selection criterion. For each candidate $K \in \{2, \ldots, K_{\max}\}$, we run $K$-means on the prediction matrices $\{\bm{p}_i\}$ and compute the Silhouette Score:
\begin{equation}
\label{eq:silhouette}
S(K) = \frac{1}{N} \sum_{i=1}^{N} \frac{b_i - a_i}{\max(a_i, b_i)},
\end{equation}
where $a_i$ is the mean distance from client $i$ to all other clients in the same cluster, and $b_i$ is the minimum mean distance from client $i$ to clients in any other cluster. The optimal cluster number is selected as $K^{*} = \argmax_{K} S(K)$. Since this procedure operates solely on the prediction matrices already computed on the server, it introduces negligible overhead and requires no additional communication. We set $K_{\max} = N - 1$ in practice. For each cluster $k$, we construct an ensemble teacher via: $F_k(\bm{x}) = \frac{1}{|\mathcal{C}_k|} \sum_{i \in \mathcal{C}_k} f_i(\bm{x})$.

\subsection{Stage 2: Online Synthesis and Bi-Level Cross-Cluster Optimization}

In Stage~2, data synthesis and bi-level optimization are interleaved within a single loop (Algorithm~\ref{alg:fedbicross}). At each iteration, the server first advances the synthesis process for all clusters, then performs bi-level updates at selected iterations. 

\subsubsection{Trajectory-Based Synthesis.}
For each cluster $k$, we generate synthetic data via DeepInversion over $T$ iterations with batch size $B$. Let $\hat{\bm{x}}_k^{(0)}$ be the initial random noise. At each iteration $t \in \{1, \ldots, T\}$, we update:
\begin{equation}
\label{eq:synthesis_update}
\hat{\bm{x}}_k^{(t)} = \hat{\bm{x}}_k^{(t-1)} - \eta_s \nabla_{\hat{\bm{x}}} \mathcal{L}_{\text{DI}}(\hat{\bm{x}}_k^{(t-1)}; F_k, y),
\end{equation}
where $\eta_s$ is the synthesis learning rate and $y$ is the target class label for the synthetic sample, which is uniformly sampled from all classes to ensure balanced generation. The DeepInversion loss~\cite{yin2020dreaming} is defined as: $\mathcal{L}_{\text{DI}}(\hat{\bm{x}}; F, y) = \mathcal{L}_{\text{CE}}(F(\hat{\bm{x}}), y) + \alpha_{\text{TV}} \mathcal{R}_{\text{TV}}(\hat{\bm{x}}) + \alpha_{\text{BN}} \mathcal{R}_{\text{BN}}(\hat{\bm{x}}),$
where $\mathcal{L}_{\text{CE}}$, $\mathcal{R}_{\text{TV}}$ and $\mathcal{R}_{\text{BN}}$ are cross-entropy, total variation, and BN losses, and $\alpha_{\text{TV}}$ and $\alpha_{\text{BN}}$ are hyper-parameters.

Rather than using only the final synthesized images, we leverage the full synthesis trajectory $\{\hat{\bm{x}}_k^{(t)}\}_{t=1}^{T}$ as training sources to augment data diversity.

\subsubsection{Noise-Adapted Teachers.}

The teacher's BN statistics are calibrated for real images, causing unreliable predictions on early-stage noisy inputs. Following FedISCA~\cite{kang2023one}, we construct a noise-adapted teacher $\tilde{F}_k$, but rather than requiring a separate post-hoc pass over the stored trajectory, we build it \emph{online} by progressively updating its BN statistics as each synthetic batch is generated. Specifically, $(\tilde{\mu}_k, \tilde{\sigma}_k^{2})$ are initialized from the original BN statistics of $F_k$. At each synthesis iteration $t$, after generating $\hat{\bm{x}}_k^{(t)}$, we update:
\begin{equation}
\label{eq:bn_update}
\tilde{\mu}_k \leftarrow \beta \tilde{\mu}_k + (1-\beta)\,\mu(\hat{\bm{x}}_k^{(t)}), \quad \tilde{\sigma}_k^{2} \leftarrow \beta \tilde{\sigma}_k^{2} + (1-\beta)\,\sigma^2(\hat{\bm{x}}_k^{(t)}),
\end{equation}
where $\beta$ is the momentum. Initializing from the original statistics ensures stability in the first few iterations (where $\beta=0.9$ keeps the statistics close to their initial values), while the progressive update gradually adapts the teacher to the evolving noise characteristics of the synthesis trajectory. This yields a noise-adapted teacher $\tilde{F}_k$ that provides reliable supervision across all noise levels.

\subsubsection{Problem Formulation.}

While clustering produces coherent teachers, intra-cluster data alone leads to overfitting. Incorporating cross-cluster data can improve diversity, but risks \textbf{negative transfer} from incompatible distributions.

The central idea behind our bi-level formulation is that \textbf{the usefulness of cross-cluster knowledge should be measured by its effect on the target cluster's performance}. This naturally leads to a bi-level optimization formulation: the \emph{inner level} trains the cluster model using weighted cross-cluster data, while the \emph{outer level} evaluates the resulting model on the target cluster's validation set and adjusts the weights accordingly. To enable this, we partition each synthetic \emph{batch} into training and validation subsets: for every batch of $B$ samples generated at any iteration $t$, a fixed index partition $\mathcal{I}^{\text{train}} \cup \mathcal{I}^{\text{val}} = \{1, \ldots, B\}$ with $\mathcal{I}^{\text{train}} \cap \mathcal{I}^{\text{val}} = \emptyset$ assigns samples to training and validation. The superscripts $\hat{\bm{x}}_k^{(t, \text{train})}$ and $\hat{\bm{x}}_k^{(t, \text{val})}$ denote these within-batch splits at iteration $t$.

Formally, for each cluster $k$, we introduce learnable cross-cluster weights $\bm{w}_k = (w_{k,1}, \ldots, w_{k,K})^\top$, where $w_{k,j} \geq 0$, $\sum_{j=1}^{K} w_{k,j} = 1$, and $w_{k,j}$ represents the contribution of cluster $j$'s knowledge to training cluster $k$'s model. Let $G_k$ denote the cluster model for cluster $k$. We formulate the following bi-level optimization problem over the entire synthesis trajectory:
\begin{equation}
\label{eq:bilevel}
\begin{aligned}
& \bm{w}_k^{*} = \argmin_{\bm{w}_k} \; \sum_{t=1}^{T} \mathcal{L}_{\text{KD}}^{(t)}\bigl(G_k^{*}(\bm{w}_k), F_k, \tilde{F}_k, \hat{\bm{x}}_k^{(t, \text{val})}\bigr) \\
& \text{s.t.} \;\; G_k^{*}(\bm{w}_k) = \argmin_{G} \sum_{t=1}^{T} \sum_{j=1}^{K} w_{k,j} \cdot \mathcal{L}_{\text{KD}}^{(t)}\bigl(G, F_j, \tilde{F}_j, \hat{\bm{x}}_j^{(t, \text{train})}\bigr),
\end{aligned}
\end{equation}
where the knowledge distillation loss is: 
\begin{equation}
\label{eq:kd_loss}
\mathcal{L}_{\text{KD}}^{(t)}(G, F, \tilde{F}, \hat{\bm{x}}) = \lambda^{(t)} \cdot \text{KL}\bigl(\tilde{F}(\hat{\bm{x}}) \,\|\, G(\hat{\bm{x}})\bigr) + (1-\lambda^{(t)}) \cdot \text{KL}\bigl(F(\hat{\bm{x}}) \,\|\, G(\hat{\bm{x}})\bigr),
\end{equation}
with $\lambda^{(t)} = 1 - t/T$. Since $\tilde{F}_k$ is initialized from $F_k$ and evolves slowly (momentum $\beta=0.9$), $\lambda^{(t)} \approx 1$ at early iterations effectively defaults to the original teacher's behavior, while at later iterations the noise-adapted statistics have stabilized and $\lambda^{(t)}$ naturally shifts reliance to $F_k$ whose BN statistics are well-calibrated for the near-realistic samples.

The \textbf{inner optimization} trains $G_k$ on a weighted mixture of all clusters' synthetic data across all iterations, where $w_{k,j}$ controls how much cluster $j$ contributes. The \textbf{outer optimization} evaluates how well this trained model performs on cluster $k$'s own validation data aggregated over the entire trajectory, and adjusts the weights to minimize the total validation loss. Through this loop, clusters providing beneficial knowledge will have their weights increased.

\subsubsection{Online Optimization.}

Directly solving the batch objective in Eq.~(\ref{eq:bilevel}) requires materializing the entire synthesis trajectory $\{\hat{\bm{x}}_k^{(t)}\}_{t=1}^{T}$ for all $K$ clusters before optimization can begin, incurring $O(T \times K \times B)$ memory. For typical settings ($T{=}500$, $K{=}5$, $B{=}256$), this amounts to several gigabytes, which is prohibitive when GPU memory must also accommodate model parameters and optimizer states.

We observe that both the inner and outer objectives in Eq.~(\ref{eq:bilevel}) decompose additively over synthesis iterations, and each term depends only on the samples at that iteration. This structure admits an \textbf{online reformulation} that interleaves synthesis and bi-level optimization within a single loop (Algorithm~\ref{alg:fedbicross}, Stage~2). At iteration $t$, we first generate the current synthetic batch $\{\hat{\bm{x}}_j^{(t)}\}_{j=1}^{K}$ via Eq.~(\ref{eq:synthesis_update}), then immediately perform the bi-level update using only the current batch:
\begin{equation}
\label{eq:online_bilevel}
\begin{aligned}
& \bm{w}_k^{(t)} = \argmin_{\bm{w}_k} \; \mathcal{L}_{\text{KD}}^{(t)}\bigl(G_k^{(t)}(\bm{w}_k), F_k, \tilde{F}_k, \hat{\bm{x}}_k^{(t, \text{val})}\bigr) \\
& \text{s.t.} \;\; G_k^{(t)}(\bm{w}_k) = \argmin_{G} \sum_{j=1}^{K} w_{k,j} \cdot \mathcal{L}_{\text{KD}}^{(t)}\bigl(G, F_j, \tilde{F}_j, \hat{\bm{x}}_j^{(t, \text{train})}\bigr),
\end{aligned}
\end{equation}
where $G_k^{(t)}$ and $\bm{w}_k^{(t)}$ are warm-started from their previous values. Since only the current batch is needed, previous synthetic data can be discarded immediately, reducing memory from $O(T \times K \times B)$ to $O(K \times B)$. We approximate Eq.~(\ref{eq:online_bilevel}) with single-step gradient updates:

\paragraph{Inner Update.} At iteration $t$, given the current weights $\bm{w}_k^{(t-1)}$, the cluster model is updated using the current synthetic batch on the training split: 
\begin{equation}
\label{eq:inner}
G_k^{(t)} = G_k^{(t-1)} - \eta_G \nabla_{G} \sum_{j=1}^{K} w_{k,j}^{(t-1)} \cdot \mathcal{L}_{\text{KD}}^{(t)}\bigl(G_k^{(t-1)}, F_j, \tilde{F}_j, \hat{\bm{x}}_j^{(t,\text{train})}\bigr),
\end{equation}
where $\eta_G$ is the learning rate for the cluster model.

\paragraph{Outer Update.} The cross-cluster weights are then adjusted based on the updated model's performance on the validation split of the same iteration:
\begin{equation}
\label{eq:outer}
\bm{w}_{k}^{(t)} = \bm{w}_{k}^{(t-1)} - \eta_w \nabla_{\bm{w}} \mathcal{L}_{\text{KD}}^{(t)}\bigl(G_k^{(t)}, F_k, \tilde{F}_k, \hat{\bm{x}}_k^{(t,\text{val})}\bigr),
\end{equation}
where $\eta_w$ is the learning rate for the weights. After each update, we project $\bm{w}_{k}^{(t)}$ onto the probability simplex to ensure $w_{k,j}^{(t)} \geq 0$ and $\sum_{j=1}^{K} w_{k,j}^{(t)} = 1$. 

By accumulating single-step updates over $t=1,\ldots,T$, the online procedure approximates the batch objective in Eq.~(\ref{eq:bilevel}) via stochastic optimization, where each iteration provides a gradient estimate from the current synthesis step.

\subsubsection{Trajectory Sampling.}

Consecutive synthesis iterations produce highly correlated samples, so not every iteration yields informative gradients for the bi-level update. We adopt a stratified sampling strategy: the trajectory $\{1, \ldots, T\}$ is divided into $S$ equal intervals and one iteration is drawn uniformly from each:
\begin{equation}
\label{eq:sampling}
\mathcal{S} = \left\{ t_s \;\middle|\; t_s \sim \mathrm{Uniform}\!\left(\left\lfloor\frac{(s-1)T}{S}\right\rfloor + 1,\; \left\lfloor\frac{sT}{S}\right\rfloor\right),\; s = 1,\ldots,S \right\}.
\end{equation}
At iterations $t \notin \mathcal{S}$, only the synthesis and BN update steps are executed and the bi-level update is skipped, reducing the number of inner/outer gradient steps from $T$ to $S$ per cluster. Since $\bm{w}_k$ has only $K$ dimensions, it converges with very few updates; we use $S{=}6$ in all experiments (see Section~\ref{sec:sampling_ablation}).

\begin{algorithm}[t]
\caption{FedBiCross}
\label{alg:fedbicross}
\KwIn{Client models $\{f_i\}_{i=1}^N$, private datasets $\{\mathcal{D}_i\}_{i=1}^N$, synthesis iterations $T$, sampling size $S$}
\KwOut{Personalized models $\{f_i^{\text{pers}}\}_{i=1}^N$}

\tcp{Stage 1: Client Clustering}
Generate random noise images $\{\bm{z}_m\}_{m=1}^M$\;
Compute prediction matrices $\bm{p}_i$ for each client $i$\;
Select $K^{*} = \argmax_{K} S(K)$ via Silhouette Score (Eq.~\ref{eq:silhouette})\;
Cluster clients into $\{\mathcal{C}_k\}_{k=1}^{K^{*}}$ via $K$-means on $\{\bm{p}_i\}$\;
Construct ensemble teachers $\{F_k\}_{k=1}^{K^{*}}$\;

\tcp{Stage 2: Online Synthesis \& Bi-Level Optimization}
Sample trajectory indices $\mathcal{S}$ via stratified sampling (Eq.~\ref{eq:sampling})\;
\For{$k = 1$ \KwTo $K^{*}$}{
    Initialize $\hat{\bm{x}}_k^{(0)} \sim \mathcal{N}(0, 1)$, noise-adapted BN $(\tilde{\mu}_k, \tilde{\sigma}_k^{2})$ from $F_k$\;
    Initialize cluster model $G_k^{(0)}$ randomly, weights $\bm{w}_k^{(0)} \leftarrow \mathbf{1}/K^{*}$\;
}
\For{$t = 1$ \KwTo $T$}{
    \For{$k = 1$ \KwTo $K^{*}$}{
        Update synthetic data: $\hat{\bm{x}}_k^{(t)} \leftarrow$ Eq.~(\ref{eq:synthesis_update})\;
        Update noise-adapted teacher: $(\tilde{\mu}_k, \tilde{\sigma}_k^{2}) \leftarrow$ Eq.~(\ref{eq:bn_update})\;
    }
    \If{$t \in \mathcal{S}$}{
        \For{$k = 1$ \KwTo $K^{*}$}{
            Inner update: $G_k^{(t)} \leftarrow$ Eq.~(\ref{eq:inner})\;
            Outer update: $\bm{w}_{k}^{(t)} \leftarrow$ Eq.~(\ref{eq:outer})\;
        }
    }
}

\tcp{Stage 3: Personalized Distillation}
\For{$k = 1$ \KwTo $K^{*}$}{
    \For{$i \in \mathcal{C}_k$}{
        Initialize $f_i^{\text{pers}} \leftarrow G_k^{(T)}$\;
        Fine-tune $f_i^{\text{pers}}$ on local data $\mathcal{D}_i$ via Eq.~(\ref{eq:pers_loss})\;
    }
}
\Return $\{f_i^{\text{pers}}\}_{i=1}^N$
\end{algorithm}

\subsection{Stage 3: Personalized Knowledge Distillation}

After $T$ iterations, we obtain cluster models $\{G_k^{(T)}\}$ that capture cluster-level knowledge through cross-cluster optimization. Since clients within the same cluster may still exhibit heterogeneous data distributions that the cluster model cannot fully capture, we further personalize it for each individual client using their private local data. For client $i$ belonging to cluster $\mathcal{C}_k$, we initialize a personalized model $f_i^{\text{pers}}$ from the cluster model $G_k^{(T)}$ and fine-tune it by minimizing the following loss function: $\mathcal{L}_{\mathrm{pers}}=$

\begin{equation}
\label{eq:pers_loss}
\mathcal{L}_{\mathrm{CE}}\bigl(f_i^{\mathrm{pers}}(\bm{x}), y\bigr) + \gamma \cdot \mathrm{KL}\bigl(G_k^{(T)}(\bm{x}) \parallel f_i^{\mathrm{pers}}(\bm{x})\bigr) + \delta \cdot \mathrm{KL}\bigl(f_i(\bm{x}) \parallel f_i^{\mathrm{pers}}(\bm{x})\bigr),
\end{equation}
where $(\bm{x}, y) \in \mathcal{D}_i$ denotes samples from client $i$'s local dataset. The three terms serve complementary purposes: (1) the cross-entropy loss $\mathcal{L}_{\text{CE}}$ trains the model to fit the client's local data distribution; (2) the KL divergence from the cluster model $G_k^{(T)}$ acts as a regularizer that preserves the generalizable knowledge learned through cross-cluster optimization, preventing catastrophic forgetting; (3) the KL divergence from the original client model $f_i$ retains the client's pre-trained local knowledge that may be lost during cluster-level distillation. 
The hyper-parameter $\delta$ controls the trade-off between retaining useful client-specific knowledge and reintroducing the non-IID bias inherent in $f_i$.

\section{Experiments}
\label{sec:experiments}

\subsection{Experimental Setup}
\subsubsection{Datasets.}
We evaluate our method on four MedMNIST datasets~\cite{medmnistv2}: BloodMNIST, DermaMNIST, OCTMNIST, and TissueMNIST. To simulate data heterogeneity, we partition the dataset using a Dirichlet distribution Dir($\alpha$) with concentration parameter $\alpha \in \{0.1, 0.2, 0.3, 0.5\}$, where lower $\alpha$ indicates higher heterogeneity. We consider three client settings: $N=5$, $N=10$, and a larger-scale setting. In the latter, we use $N=20$ clients for all datasets except DermaMNIST, which uses $N=15$ due to its smaller sample size. For each setting, the number of clusters $K$ is automatically determined by the Silhouette Score (Eq.~\ref{eq:silhouette}). To provide a comprehensive analysis, we also report results with alternative $K$ values to demonstrate the effectiveness of the automatic selection.

\subsubsection{Baselines.}
We compare against five one-shot FL baselines: FedAvg with a single communication round (FedAvg-1)~\cite{mcmahan2017communication}, DAFL~\cite{chen2019data}, DENSE~\cite{zhang2022dense}, FedISCA~\cite{kang2023one}, and Co-Boosting~\cite{dai2024enhancing}. All methods are evaluated by test accuracy on each client, and we report the averaged accuracy across clients.

\subsubsection{Implementation Details.}
We adopt ResNet-18 as the backbone. Client models are trained for 100 epochs using SGD with learning rate $10^{-3}$ and batch size 128. For client clustering, we use $M=256$ random noise images to compute prediction matrices, and evaluate $K \in \{2, \ldots, N-1\}$ using the Silhouette Score to automatically select the optimal $K^{*}$. For synthetic data generation, we optimize inputs for $T=500$ iterations with batch size $B=256$ using Adam with learning rate $5\times10^{-2}$, where 80\% of samples are used for training and 20\% for validation in bi-level optimization. We apply stratified trajectory sampling with $S=6$, which reduces the number of bi-level update steps from $T=500$ to only 6 while maintaining performance (see Section~\ref{sec:sampling_ablation}). We set $\alpha_{\mathrm{TV}}=2.5\times10^{-5}$, $\alpha_{\mathrm{BN}}=10$, temperature $\tau=20$, and momentum $\beta=0.9$ for noise adaptation. For personalized distillation, we set $\gamma=0.5$, $\delta=0.3$ and fine-tune for 10 epochs.

\subsection{Main Results}
\begin{table}[!ht]
\centering
\caption{Comparison of different methods across client counts and heterogeneity levels (Accuracy\%). For $N=20$, DermaMNIST uses $N=15$ due to its smaller sample size. \textbf{Bold} entries indicate the $K$ automatically selected.}
\label{tab:main}
\resizebox{\textwidth}{!}{
\begin{tabular}{cl|cccc|cccc|cccc|cccc}
\toprule
& & \multicolumn{4}{c|}{BloodMNIST} & \multicolumn{4}{c|}{DermaMNIST} & \multicolumn{4}{c|}{OCTMNIST} & \multicolumn{4}{c}{TissueMNIST} \\
\cmidrule(lr){3-6}\cmidrule(lr){7-10}\cmidrule(lr){11-14}\cmidrule(lr){15-18}
$N$ & Method & 0.1 & 0.2 & 0.3 & 0.5 & 0.1 & 0.2 & 0.3 & 0.5 & 0.1 & 0.2 & 0.3 & 0.5 & 0.1 & 0.2 & 0.3 & 0.5 \\
\midrule
\multirow{8}{*}{5}
& FedAvg-1     & 14.23 & 16.37 & 17.35 & 18.26 & 15.62 & 16.48 & 17.24 & 64.62 & 28.86 & 30.34 & 31.10 & 33.75 & 29.78 & 31.45 & 33.76 & 34.08 \\
& DAFL         & 13.42 & 14.05 & 15.74 & 17.03 & 15.96 & 16.86 & 17.71 & 66.31 & 27.39 & 29.12 & 31.58 & 36.74 & 29.33 & 31.56 & 35.01 & 45.78 \\
& DENSE        & 42.34 & 46.78 & 50.45 & 75.78 & 16.85 & 17.60 & 18.45 & 66.56 & 49.72 & 51.46 & 54.32 & 61.83 & 28.78 & 31.71 & 33.88 & 40.63 \\
& FedISCA      & 48.49 & 50.64 & 53.28 & 79.83 & 18.12 & 19.89 & 21.84 & 68.05 & 52.05 & 56.26 & 60.47 & 67.58 & 43.81 & 44.98 & 49.86 & 52.15 \\
& Co-Boosting  & 54.75 & 56.16 & 58.41 & 80.32 & 38.29 & 43.55 & 46.08 & 67.40 & 51.78 & 54.90 & 57.63 & 65.77 & 48.92 & 51.03 & 53.26 & 56.03 \\
\cmidrule(l){2-18}
& Ours ($K$=2) & 84.94 & 86.51 & 87.15 & 88.65 & 65.17 & 67.03 & 68.12 & \textbf{73.21} & 61.95 & 63.78 & 66.62 & \textbf{71.36} & 61.28 & 62.75 & 63.71 & \textbf{65.02} \\
& Ours ($K$=3) & 83.31 & 83.90 & \textbf{87.67} & \textbf{90.13} & \textbf{65.82} & \textbf{69.66} & \textbf{71.45} & 71.29 & \textbf{63.77} & \textbf{65.42} & \textbf{68.01} & 70.12 & 60.03 & \textbf{63.97} & \textbf{64.81} & 64.21 \\
& Ours ($K$=4) & \textbf{85.57} & \textbf{87.12} & 85.63 & 86.17 & 64.52 & 68.40 & 67.22 & 71.18 & 60.86 & 64.14 & 67.23 & 69.88 & \textbf{62.54} & 62.52 & 64.60 & 64.97 \\
\midrule
\multirow{8}{*}{10}
& FedAvg-1     & 13.58 & 14.95 & 16.28 & 17.41 & 14.92 & 15.86 & 16.44 & 61.85 & 27.84 & 29.41 & 30.22 & 32.96 & 28.66 & 30.78 & 33.01 & 33.54 \\
& DAFL         & 12.74 & 13.31 & 14.73 & 16.07 & 15.18 & 15.96 & 17.74 & 60.62 & 26.52 & 27.98 & 29.57 & 33.88 & 26.89 & 30.75 & 32.29 & 41.97 \\
& DENSE        & 38.62 & 40.73 & 45.12 & 64.45 & 16.09 & 16.91 & 17.63 & 63.18 & 48.97 & 50.55 & 53.46 & 57.92 & 27.12 & 29.84 & 33.02 & 37.72 \\
& FedISCA      & 44.28 & 45.97 & 48.60 & 74.31 & 17.43 & 19.03 & 20.88 & 64.63 & 51.28 & 54.66 & 57.34 & 63.20 & 39.76 & 42.04 & 47.38 & 51.87 \\
& Co-Boosting  & 49.52 & 52.67 & 54.10 & 72.89 & 35.63 & 40.76 & 43.21 & 63.28 & 50.67 & 53.82 & 55.38 & 61.04 & 43.88 & 46.63 & 49.94 & 54.28 \\
\cmidrule(l){2-18}
& Ours ($K$=3) & 77.59 & 80.37 & 81.61 & \textbf{86.55} & 62.18 & 64.48 & 68.29 & \textbf{71.08} & 61.18 & 63.56 & \textbf{67.69} & \textbf{69.11} & 57.13 & 59.05 & 61.37 & \textbf{63.38} \\
& Ours ($K$=4) & 78.67 & \textbf{82.04} & \textbf{84.13} & 84.26 & \textbf{63.84} & \textbf{66.20} & \textbf{69.93} & 70.26 & \textbf{62.68} & \textbf{65.04} & 66.81 & 65.89 & 58.79 & 60.57 & \textbf{62.73} & 62.10 \\
& Ours ($K$=5) & \textbf{79.67} & 81.90 & 82.04 & 85.60 & 61.01 & 65.20 & 69.11 & 69.97 & 62.11 & 63.47 & 66.83 & 67.36 & \textbf{59.11} & \textbf{61.25} & 61.82 & 62.45 \\
\midrule
\multirow{8}{*}{20}
& FedAvg-1     & 12.95 & 13.70 & 14.01 & 15.28 & 14.76 & 15.23 & 14.85 & 58.44 & 25.98 & 27.65 & 27.81 & 29.40 & 22.43 & 24.82 & 26.19 & 29.24 \\
& DAFL         & 12.55 & 12.83 & 13.17 & 14.29 & 14.55 & 14.96 & 16.57 & 57.40 & 25.43 & 26.97 & 28.22 & 31.58 & 20.96 & 23.77 & 25.49 & 36.86 \\
& DENSE        & 29.90 & 32.46 & 34.16 & 65.85 & 15.24 & 15.89 & 16.33 & 58.25 & 46.89 & 48.11 & 50.57 & 55.98 & 22.90 & 24.66 & 27.92 & 30.85 \\
& FedISCA      & 31.47 & 36.50 & 42.89 & 72.52 & 17.02 & 17.85 & 18.91 & 60.42 & 49.30 & 52.84 & 53.27 & 58.75 & 32.56 & 35.87 & 41.96 & 48.24 \\
& Co-Boosting  & 37.09 & 43.81 & 48.97 & 70.35 & 31.22 & 35.28 & 38.60 & 60.59 & 47.81 & 51.68 & 52.84 & 56.53 & 37.43 & 38.10 & 43.48 & 49.78 \\
\cmidrule(l){2-18}
& Ours ($K$=3) & 64.65 & 67.18 & 68.34 & \textbf{80.08} & 51.14 & 53.46 & \textbf{58.46} & \textbf{67.10} & 59.82 & 62.96 & 65.68 & \textbf{68.14} & 52.22 & 53.89 & 56.74 & \textbf{59.56} \\
& Ours ($K$=5) & 64.88 & \textbf{68.24} & \textbf{70.49} & 77.92 & \textbf{52.61} & \textbf{55.85} & 57.07 & 65.58 & \textbf{61.26} & \textbf{64.33} & \textbf{67.07} & 67.71 & 52.86 & \textbf{55.21} & \textbf{57.09} & 59.19 \\
& Ours ($K$=8) & \textbf{66.73} & 65.41 & 68.71 & 76.50 & 50.73 & 52.98 & 56.62 & 65.03 & 59.88 & 63.40 & 66.11 & 67.68 & \textbf{53.68} & 54.09 & 55.31 & 58.70 \\
\bottomrule
\end{tabular}
}
\end{table}

Table~\ref{tab:main} summarizes results on four MedMNIST datasets under different client counts and heterogeneity levels. The \textbf{bold} entries correspond to the $K$ value automatically selected by the Silhouette Score (Eq.~\ref{eq:silhouette}); results with alternative $K$ values are also shown for reference. FedBiCross with the auto-selected $K$ achieves the best or near-best accuracy in the vast majority of settings, and yields substantial gains over the strongest baseline in all cases.
 
On BloodMNIST with $N=5$ and $\alpha=0.1$, the auto-selected $K^{*}=4$ surpasses Co-Boosting by over 30 points. The auto-selected $K$ consistently matches or closely approaches the best $K$ across most dataset and heterogeneity combinations. For instance, with $N=10$ on DermaMNIST ($\alpha=0.1$), the auto-selected $K^{*}=4$ outperforms all alternative $K$ values, validating that the Silhouette Score effectively identifies the cluster granularity that balances intra-cluster coherence with sufficient per-cluster data.
 
Under increased heterogeneity (as $\alpha$ drops from 0.5 to 0.1), Co-Boosting drops by over 25 points on BloodMNIST ($N=5$), while FedBiCross degrades by less than 5 points, demonstrating strong robustness. The trend persists in larger client settings: with $N=20$ and $\alpha=0.1$, FedBiCross outperforms Co-Boosting by nearly 30 points.
 
We also observe that the auto-selected $K$ tends to increase with the number of clients $N$: for $N=5$, $K^{*}$ is typically 2--4; for $N=10$, $K^{*}$ is 3--5; and for $N=20$, $K^{*}$ reaches 3--8. This is intuitive, as more clients allow finer-grained clustering while maintaining sufficient members per cluster. Under high heterogeneity ($\alpha=0.1$), the Silhouette Score favors larger $K$ to ensure intra-cluster consistency, whereas under mild heterogeneity ($\alpha=0.5$), smaller $K$ is preferred to preserve data sufficiency per cluster.

\subsection{Ablation Study}
The ablations in Table~\ref{tab:ablation} verify the effect of each component. For cross-cluster knowledge utilization, bi-level optimization consistently outperforms all alternatives. On BloodMNIST with $N=5$ and $\alpha=0.1$, the gap from intra-cluster only to the full method exceeds 5 points, with uniform and similarity-weighted cross-cluster schemes falling in between, showing that adaptive weighting surpasses heuristic designs. Removing personalized distillation (w/o PKD) yields large drops across all settings, e.g., over 12 points on DermaMNIST with $N=10$ and $\alpha=0.1$, indicating that cluster models alone fail to capture client-specific patterns. Removing clustering (w/o Clus) degrades performance further, confirming that clustering is essential for forming effective teachers.
\begin{table}[!t]
\centering
\caption{Ablation study across different client counts and heterogeneity levels (Accuracy\%). All variants use the auto-selected $K$; w/o Clus uses $K=1$ (i.e., all clients form a single group).}
\label{tab:ablation}
\resizebox{\textwidth}{!}{
\begin{tabular}{cl|cccc|cccc|cccc|cccc}
\toprule
& & \multicolumn{4}{c|}{BloodMNIST} & \multicolumn{4}{c|}{DermaMNIST} & \multicolumn{4}{c|}{OCTMNIST} & \multicolumn{4}{c}{TissueMNIST} \\
\cmidrule(lr){3-6}\cmidrule(lr){7-10}\cmidrule(lr){11-14}\cmidrule(lr){15-18}
$N$ & Variant & 0.1 & 0.2 & 0.3 & 0.5 & 0.1 & 0.2 & 0.3 & 0.5 & 0.1 & 0.2 & 0.3 & 0.5 & 0.1 & 0.2 & 0.3 & 0.5 \\
\midrule
\multirow{5}{*}{5}
& Intra-cluster & 80.01 & 81.75 & 83.07 & 85.92 & 60.08 & 64.11 & 66.12 & 69.10 & 56.56 & 61.78 & 65.56 & 68.22 & 60.28 & 61.99 & 63.90 & 63.98 \\
& Uniform cross & 81.66 & 83.35 & 84.44 & 86.17 & 61.79 & 65.76 & 67.30 & 70.13 & 58.71 & 63.46 & 65.98 & 69.06 & 60.95 & 62.58 & 64.17 & 64.21 \\
& Sim-weighted  & 82.99 & 83.63 & 85.13 & 87.02 & 63.05 & 66.08 & 67.44 & 71.16 & 59.42 & 63.80 & 66.41 & 69.64 & 61.49 & 62.64 & 64.39 & 64.42 \\
& w/o PKD       & 77.69 & 79.52 & 81.15 & 84.16 & 57.69 & 61.80 & 65.31 & 73.05 & 53.56 & 59.43 & 63.12 & 68.33 & 59.34 & 61.17 & 63.52 & 64.39 \\
& w/o Clus      & 76.47 & 76.41 & 78.40 & 82.64 & 51.28 & 56.64 & 59.35 & 69.09 & 53.03 & 58.32 & 62.19 & 68.07 & 56.61 & 58.50 & 59.74 & 60.11 \\
\midrule
\multirow{5}{*}{10}
& Intra-cluster & 72.91 & 75.41 & 77.93 & 80.56 & 59.41 & 62.26 & 65.35 & 67.40 & 58.29 & 59.72 & 64.40 & 63.76 & 53.72 & 56.18 & 58.49 & 60.17 \\
& Uniform cross & 75.14 & 77.52 & 79.81 & 82.18 & 60.88 & 62.75 & 66.97 & 68.15 & 59.41 & 60.90 & 65.31 & 64.23 & 55.45 & 57.91 & 60.07 & 60.79 \\
& Sim-weighted  & 76.23 & 78.56 & 80.85 & 83.45 & 60.92 & 64.10 & 68.01 & 69.03 & 60.88 & 60.94 & 66.86 & 65.04 & 56.72 & 58.04 & 60.22 & 61.83 \\
& w/o PKD       & 67.26 & 71.26 & 74.48 & 75.89 & 51.03 & 55.63 & 59.87 & 66.68 & 54.14 & 57.87 & 61.62 & 64.09 & 53.51 & 57.28 & 59.66 & 61.31 \\
& w/o Clus      & 65.48 & 70.41 & 73.08 & 74.76 & 50.92 & 53.71 & 56.89 & 65.21 & 53.95 & 58.07 & 59.71 & 65.95 & 51.89 & 54.14 & 57.13 & 60.28 \\
\midrule
\multirow{5}{*}{20}
& Intra-cluster & 61.21 & 63.54 & 65.14 & 73.74 & 47.12 & 49.98 & 53.83 & 63.57 & 57.84 & 59.63 & 63.61 & 64.98 & 49.61 & 51.22 & 54.68 & 58.12 \\
& Uniform cross & 61.49 & 66.02 & 65.33 & 74.88 & 48.36 & 49.11 & 55.06 & 64.14 & 57.08 & 60.92 & 63.96 & 65.11 & 50.83 & 52.44 & 55.23 & 58.33 \\
& Sim-weighted  & 63.86 & 66.90 & 67.71 & 76.32 & 50.12 & 52.42 & 55.71 & 64.62 & 59.46 & 62.37 & 65.21 & 67.08 & 52.17 & 52.61 & 55.88 & 58.52 \\
& w/o PKD       & 54.62 & 58.85 & 61.47 & 72.89 & 41.32 & 46.61 & 49.45 & 63.93 & 51.18 & 54.25 & 56.32 & 59.91 & 46.24 & 49.63 & 54.07 & 57.90 \\
& w/o Clus      & 53.73 & 58.28 & 62.14 & 73.64 & 43.18 & 46.77 & 51.06 & 64.73 & 50.67 & 55.76 & 57.08 & 60.36 & 48.72 & 50.89 & 53.96 & 56.74 \\
\bottomrule
\end{tabular}
}
\end{table}

\begin{figure}[!ht]
\centerline{\includegraphics[width=0.8\textwidth]{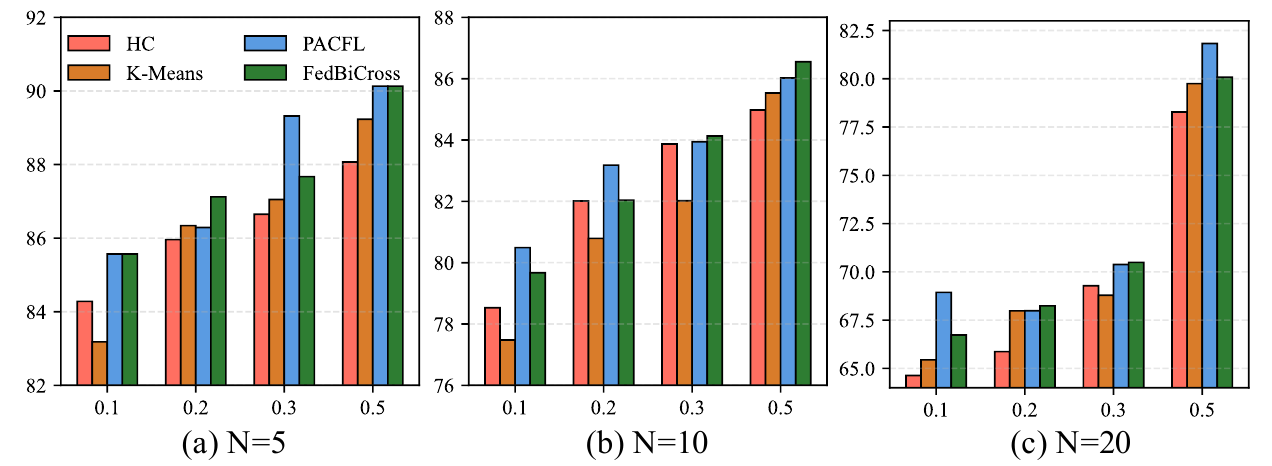}}
\caption{Ablation on the clustering stage for BloodMNIST. Test accuracy (\%) under varying client numbers $N$ and heterogeneity levels $\alpha$.}
\label{cluster}
\end{figure}

We also ablate clustering in Fig.~\ref{cluster}. Ghosh et al.~\cite{ghosh2019robust} (``$K$-means'') cluster model parameters via $K$-means; Briggs et al.~\cite{briggs2020federated} (``HC'') use hierarchical clustering. PACFL~\cite{vahidian2023efficient} clusters clients using subspaces from local data via truncated SVD, requiring data-dependent transmission and increasing privacy risk. FedBiCross instead clusters clients by prediction similarity using uploaded models alone, without extra transmission, while achieving comparable performance.

\begin{figure}[!ht]
\centerline{\includegraphics[width=0.8\textwidth]{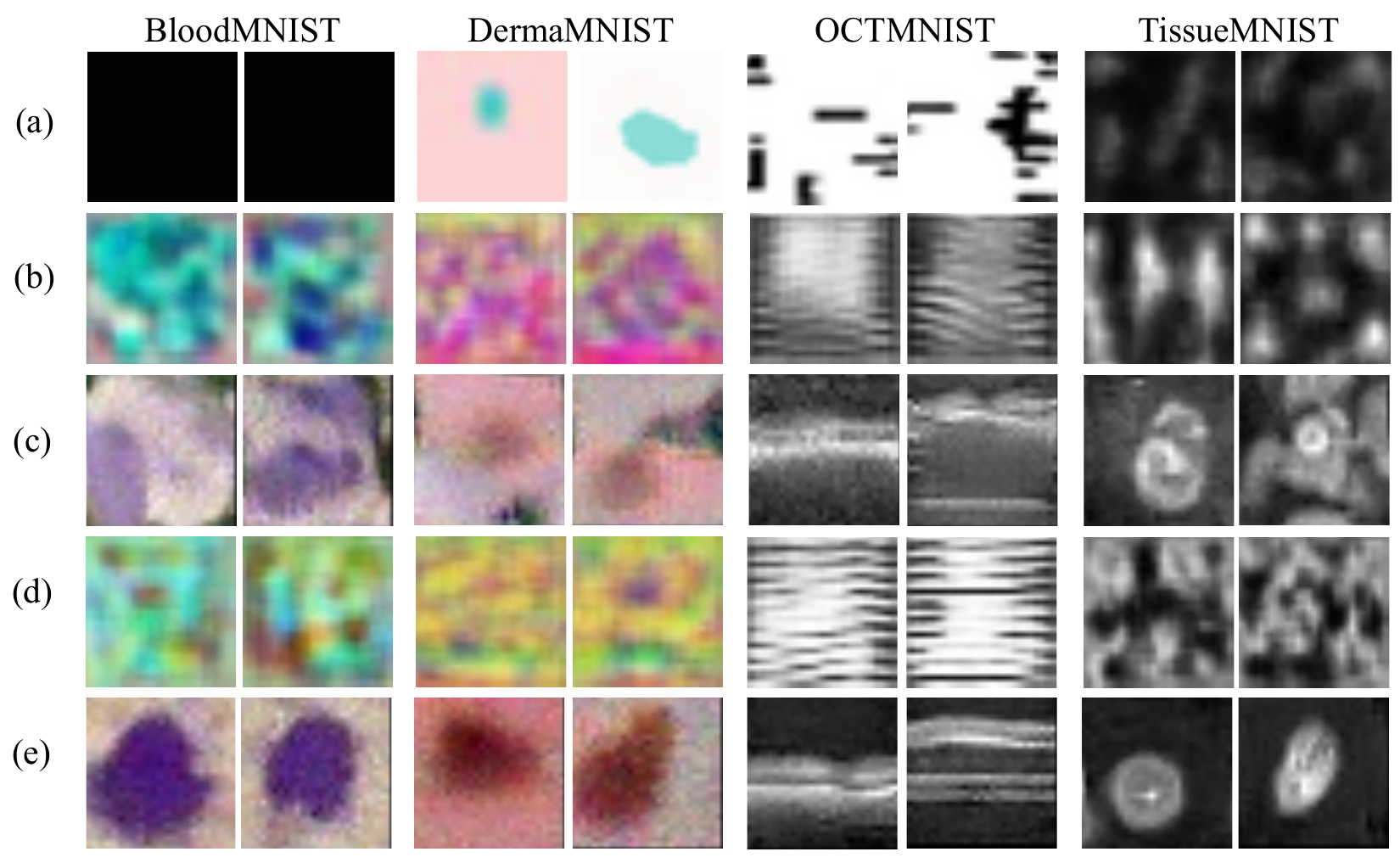}}
\caption{Visualization of synthetic images generated by different methods under non-IID settings with $\alpha=0.3$ and $N=10$: (a) DAFL, (b) DENSE, (c) FedISCA, (d) Co-Boosting, and (e) Ours.}
\label{pic_comparison}
\end{figure}

\subsection{Synthetic Data Quality}
Figure~\ref{pic_comparison} visualizes synthetic images generated by different methods under non-IID settings. DAFL suffers from severe mode collapse, producing nearly blank images on BloodMNIST and featureless color patches on DermaMNIST. DENSE generates blurry images lacking clear structure, with colors and textures that bear little resemblance to real medical images. FedISCA shows improved structure but exhibits prominent stripe artifacts, particularly on OCTMNIST, where the layered retinal structure is replaced by repetitive horizontal patterns. Co-Boosting produces images with chaotic patterns and unnatural colors, suggesting that its client-wise reweighting alone cannot resolve conflicting gradients from heterogeneous teachers. In contrast, our cluster-specific generation yields more recognizable medical features: cell boundaries and cytoplasmic textures are visible in BloodMNIST and TissueMNIST, and layered retinal structures are preserved in OCTMNIST. This improvement stems from the coherent sub-ensembles, which produce consistent gradients during DeepInversion and guide the synthesis toward realistic samples rather than averaging out conflicting signals.

\subsection{Effect of Trajectory Sampling Size}
\label{sec:sampling_ablation}
\begin{figure}[!ht]
\centerline{\includegraphics[width=0.85\textwidth]{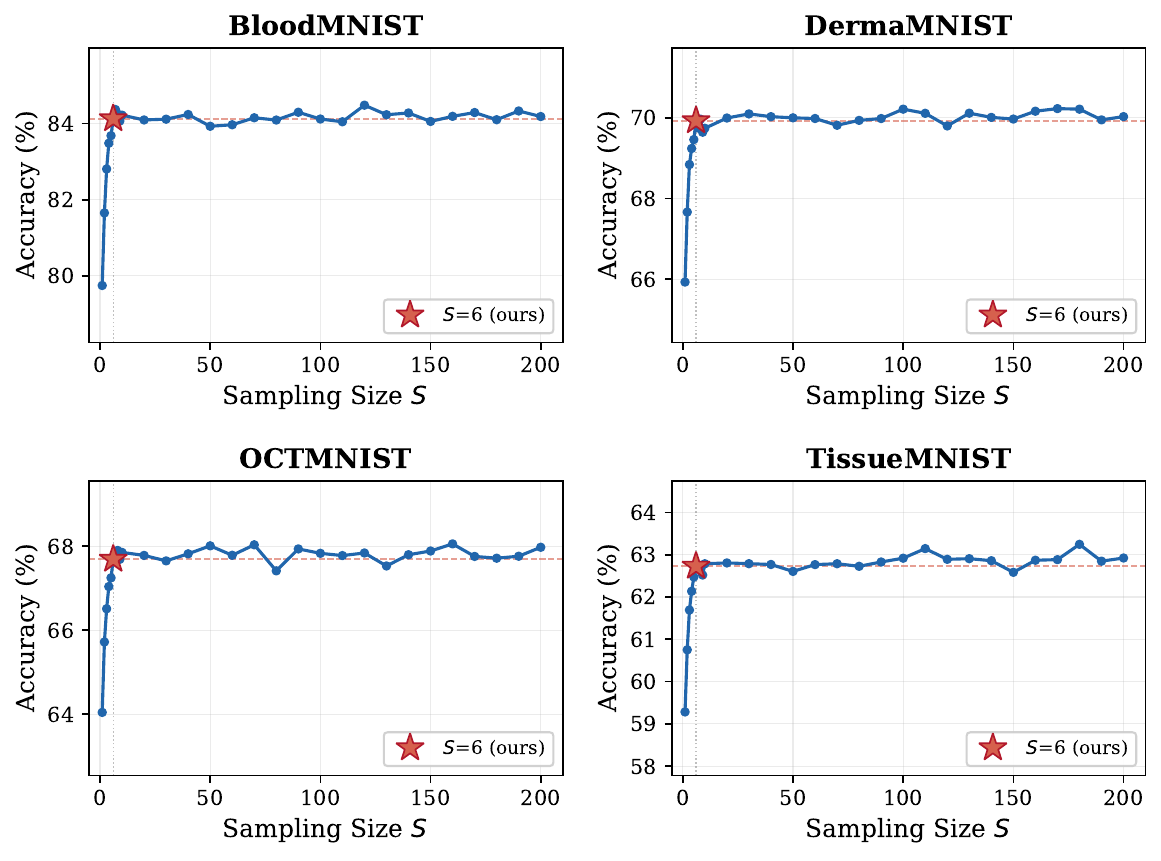}}
\caption{Effect of trajectory sampling size $S$ on test accuracy (\%) with $N=10$ and $\alpha=0.3$. The red star marks $S=6$ used in our main experiments. }
\label{fig:sampling}
\end{figure}

Figure~\ref{fig:sampling} shows the effect of trajectory sampling size $S$ on all four datasets ($N=10$, $\alpha=0.3$). At $S=1$, accuracy drops by 2\%--4\% across all datasets, confirming that a single sample is insufficient to capture the variation across noise levels. Accuracy rises sharply from $S=1$ to $S=6$ and then plateaus with marginal fluctuations ($<$0.5\%), because the low-dimensional weight vector $\bm{w}_k$ ($K$ parameters) converges in very few steps and stratified sampling ensures each step covers a distinct noise level. This pattern is consistent across all four datasets despite their different visual characteristics and class counts, suggesting that the rapid convergence is a property of the bi-level formulation rather than dataset-dependent. We set $S=6$ in all experiments, achieving 98.8\%--99.9\% of fully-sampled performance ($S=200$) at 3\% of the bi-level computation cost.

\section{Conclusion}
We presented FedBiCross, a personalized one-shot federated learning framework for non-IID medical images. By identifying that conflicting client predictions weaken soft labels under heterogeneous distributions, we proposed client clustering based on model output similarity to form coherent sub-ensembles, bi-level optimization to learn adaptive cross-cluster weights, and personalized distillation to adapt cluster-level models to individual clients. Experiments on four medical imaging datasets validated that FedBiCross consistently outperforms existing methods across different non-IID degrees, demonstrating that addressing teacher conflicts through clustering and controlled cross-cluster knowledge utilization is essential for OSFL on medical images.

\bibliographystyle{spmpsci_unsrt}
\bibliography{ref}

\end{document}